\documentclass[10pt,twocolumn,letterpaper]{article}

\usepackage{iccv}
\usepackage{times}
\usepackage{epsfig}
\usepackage{graphicx}
\usepackage{amsmath}
\usepackage{amssymb}
\usepackage{booktabs}
\usepackage{tabularx}
\usepackage{adjustbox}
\usepackage{multirow}
\usepackage{caption}
\usepackage{sidecap}
\usepackage[accsupp]{axessibility}  

\usepackage{subcaption}

\usepackage[breaklinks=true,bookmarks=false]{hyperref}

\usepackage[capitalize]{cleveref}
\crefname{section}{Sec.}{Secs.}
\Crefname{section}{Section}{Sections}
\Crefname{table}{Table}{Tables}
\crefname{table}{Tab.}{Tabs.}

\iccvfinalcopy

\def\iccvPaperID{0}

\ificcvfinal\fi

\newcommand{\methodName}{\textsc{SurfsUp}\xspace}

\newcommand{\methodNameFull}{
\textbf{Surf}ace implicit \textbf{sU}bstitution for \textbf{p}articles
}

\usepackage{stfloats}
\usepackage[dvipsnames]{xcolor}
\hypersetup{
    colorlinks,
    linkcolor={red!50!black},
    citecolor={blue!50!black},
    urlcolor={blue!80!black}
}
\newif\ifcomments
\commentstrue
\ifcomments
  \newcommand{\colornote}[3]{{\color{#1}\bf{#2: #3}\normalfont}}
\else
  \newcommand{\colornote}[3]{}
\fi

\newcommand{\printfnsymbol}[1]{%
  \textsuperscript{\@fnsymbol{#1}}
}

\begin{document}

\title{{\includegraphics[height=1.3\fontcharht\font`\B]{surfing.png}}
\methodName: Learning Fluid Simulation for Any Implicit Surface
}
\title{
\methodName: Learning Fluid Simulation for Novel Surfaces
}

\author{Arjun Mani\textsuperscript{*}\textsuperscript{1}, Ishaan Preetam Chandratreya\textsuperscript{*\textsuperscript{1}}, Elliot Creager\textsuperscript{2} \textsuperscript{3}, Carl Vondrick\textsuperscript{1}, Richard Zemel\textsuperscript{1} \\\textsuperscript{1} Columbia University \hspace{0.16cm} 
\textsuperscript{2} University of Toronto \hspace{0.16cm} \textsuperscript{3}  Vector Institute \\\href{http://surfsup.cs.columbia.edu}{\texttt{surfsup.cs.columbia.edu}}}
\twocolumn[{%
\renewcommand\twocolumn[1][]{#1}%
\maketitle

\vspace{-3mm}

\begin{center}
\vspace{-0.5cm}\includegraphics[width=0.91\linewidth]{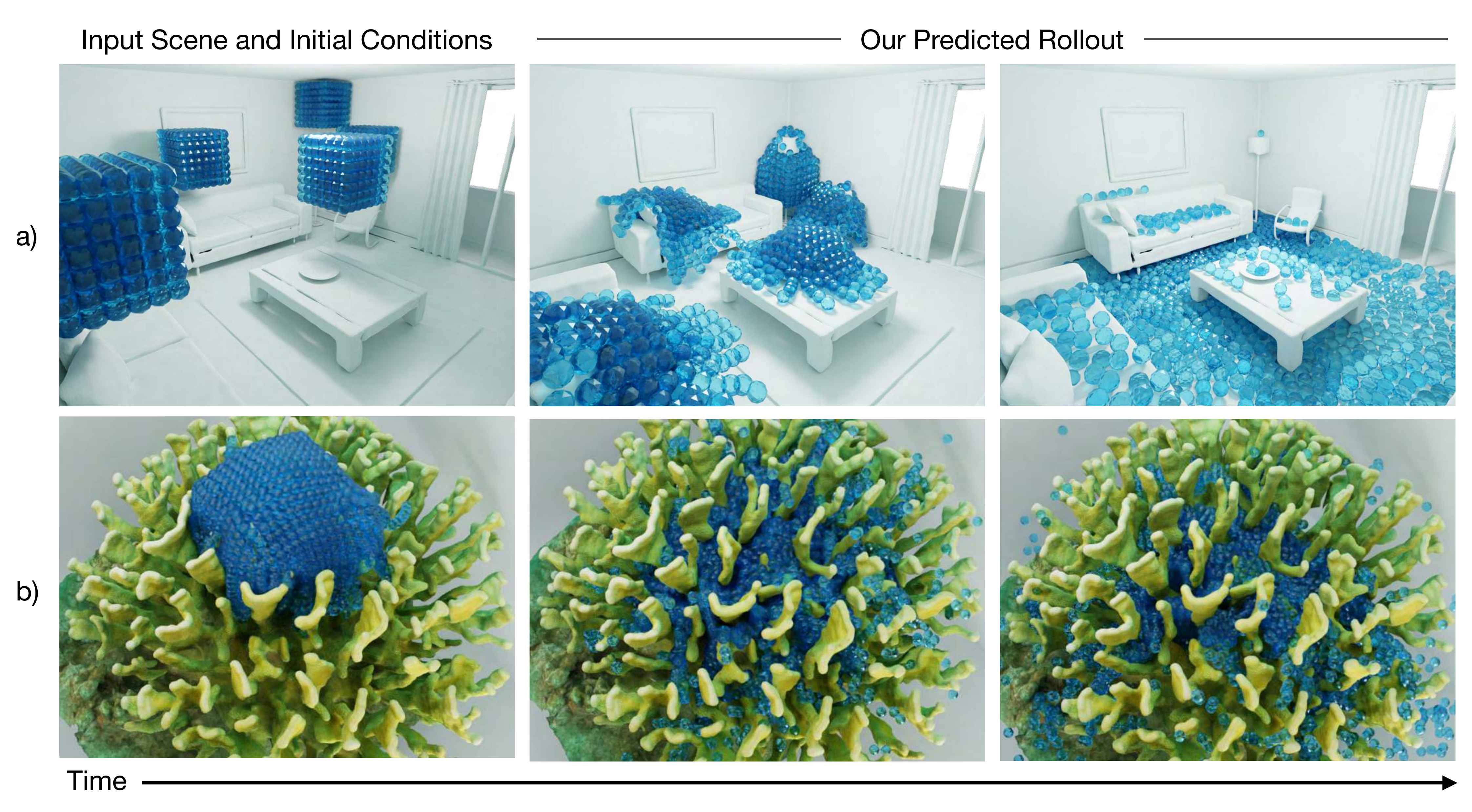}
\captionof{figure}{
We introduce \methodName, which learns to predict how fluid particles interact with novel, complex 3D surfaces. 
}
\label{fig:teaser}
\end{center}
}]
{
  \renewcommand{\thefootnote}
    {\fnsymbol{footnote}}
  \footnotetext[1]{Authors contributed equally.}
}

\ificcvfinal\thispagestyle{empty}\fi

\begin{abstract}

\vspace{-3mm}

Modeling the mechanics of fluid in complex scenes is vital to applications in design, graphics, and robotics. 
Learning-based methods provide fast and differentiable fluid simulators, however most prior work is unable to accurately model how fluids interact with genuinely novel surfaces not seen during training.
We introduce \methodName, a framework that represents objects \emph{implicitly} using signed distance functions (SDFs), rather than an explicit representation of meshes or particles.
This continuous representation of geometry enables more accurate simulation of fluid-object interactions over long time periods while simultaneously making computation more efficient. Moreover, \methodName trained on simple shape primitives generalizes considerably out-of-distribution, even to complex real-world scenes and objects.
Finally, we show we can invert our model to design simple objects to manipulate fluid flow. 
\end{abstract}

\vspace{-1.5em}
\section{Introduction}
\label{sec:intro}

Across engineering and science, the simulation of fluid dynamics has become an invaluable tool, and it will be a crucial component for building visual systems that are capable of understanding and interacting with real-world environments. Recently, a new class of simulators has emerged that \emph{learn} the dynamics of physical systems from data~\cite{li2019propagation,ummenhofer2019lagrangian,sanchez2020learning}. These simulators offer several advantages over classical simulators
by increasing speed, reducing reliance on hand-crafted dynamics models, and providing differentiable rollouts for solving inverse problems.
Recent approaches based on graph neural networks (GNNs) have in particular shown impressive accuracy and generalization on a wide range of fluids represented as particle or mesh-based systems~\cite{sanchez2020learning,pfaff2020learning}.

In a wide variety of applications areas, fluid simulation must properly handle \emph{interaction with solid structures} to attain practical relevance.
Many problems involving fluid simulation are fundamentally about object design; for example, designing aerodynamic shapes for cars and airplanes. In robotics too, an intelligent robot would have to understand fluid-surface interactions; for example, to carry a mug of hot coffee to a person without spilling the liquid from the container. In both cases, the need to accurately model how fluids interact with \emph{novel} solid surfaces\footnote{For brevity we refer to surfaces of solid objects simply as ``surfaces''.}--- with new shapes, configurations, orientations, and so on---motivates an challenging out-of-distribution (OOD) generalization subproblem for fluid simulation.

In this paper, we  tackle this problem by introducing \methodNameFull (\methodName), a framework for 
simulating the interactions of fluids and surfaces by representing surfaces using \emph{implicit representations}~\cite{michalkiewicz2019implicit}.
Current GNN-based approaches represent surfaces explicitly, discretizing them into particles and augmenting the graph accordingly~\cite{sanchez2020learning}. Implicit representations, in particular signed distance functions which represent surfaces as zero-level sets of functions, offer several compelling advantages. They can smoothly represent continuous surfaces, provide rich geometric information (e.g. whether a point is in the interior/exterior of the object), and can scale easily to large objects and scenes. Due to these properties, many state-of-the-art generative models for 3D shapes learn shapes with signed distance functions. We argue that these properties are useful not just for statics but also \emph{dynamics}:
in order to accurately model fluid-surface interactions for potentially complex surfaces, smooth and informative representations of \emph{local} surface geometry are needed.
In turn, we posit that these locally informative representations will enable fluid simulators to appropriately model interaction with never-before-seen surfaces. 

Concretely, we
propose an approach that integrates signed distance representations of surfaces~\cite{park2019deepsdf,michalkiewicz2019implicit} with graph neural network models~\cite{scarselli2008graph,sanchez2020learning}. Our architecture follows the encode-process-decode 
paradigm~\cite{sanchez2020learning}, where the particles in the physical system at time $T$ are encoded into a graph, followed by several message-passing steps, and finally a decoder predicts the dynamics (accelerations). We represent the surface using its signed distance function, and exploit geometric properties of SDFs to model the distance and orientation of fluid particles with respect to the surface.

Experiments show that our approach can model fluid-surface interactions over long rollouts with a high degree of accuracy and visual realism.
\methodName can model surfaces far beyond its training domain; when trained on simple primitive shapes (e.g. spheres, cones, toruses), we generalize \emph{zero-shot} to complex real-world objects and scenes represented using neural implicit generative models (\cref{fig:teaser}). We compare to a method which represents the surface as particles inside the graph network simulator,
and show that our model achieves better generalization and efficiency, and can better handle 
complex
surface geometry. Finally, we demonstrate the potential of 
\methodName
to solve inverse problems across two object design tasks.
Overall, our paper bridges learned physical models for dynamics and 3D implicit representations, and takes a step towards AI models which can understand and help design the physical world.

\section{Related Work}

\noindent \textbf{Learning Differentiable Simulators}. Using machine learning for predicting and modeling complex physical systems is a rapidly growing area of research. Learned physics simulators have been shown to accurately simulate colliding rigid objects, deformable objects, fluids, and other complex systems \cite{li2019propagation,battaglia2016interaction,li2018learning,ummenhofer2019lagrangian,sanchez2020learning,mrowca2018flexible}.

Recent progress in learned fluid simulation has been driven by the Lagrangian viewpoint, which tracks the motion of fluid particles over time. Graph neural networks (GNNs) have in particular emerged as effective forward models for fluid simulation, and have demonstrated better stability and generalization compared to grid-based, convolutional approaches. \cite{li2019propagation,liu2018physical,spnets2018,sanchez2020learning}. 
Sanchez-Gonzalez et al.~\cite{sanchez2020learning} introduced a GNN architecture which can accurately and generalizably model fluid behavior over long rollouts; most subsequent work builds off their approach \cite{pfaff2020learning, mayr2021boundary}. However, most of these works generally consider simple surfaces; \eg only cube-shaped containers for fluids. There has been little work focused on fluid-surface interactions; Mayr et al.~\cite{mayr2021boundary} introduce an approach specifically for triangularized meshes. All these approaches use particles or meshes to represent surfaces; we will argue that implicit representations enable learning more accurate, generalizable physical dynamics.

Some works have explored the possibility of using learned fluid simulators for solving inverse problems \cite{spnets2018,li2018learning,allen2022physical}. Allen et al.~\cite{allen2022physical} showed that graph network simulators can be used to solve design tasks by rolling out fluid trajectories and using gradient-based optimization.  In 3D, they show that a 2D plane with a learned height field can be used to split a stream of fluid to fall into several locations on the ground. SDFs have the advantage of being able to learn geometry in a more smooth and unconstrained way, which could help solve inverse problems.

\noindent \textbf{3D Implicit Representations}. Implicit representations have become state-of-the-art for generative models of 3D objects \cite{mescheder2019occupancy,park2019deepsdf,chaudhuri2020learning,sitzmann2019metasdf,sitzmann2019siren}. These models represent the surface as a function $F(x)$. There are several choices for $F$. Park et al.~\cite{park2019deepsdf} show that signed distance functions $F(x)$ are particularly effective and informative representations. Compared to point clouds \cite{achlioptas2017latent_pc,pointcloud2} or voxels \cite{voxel1,voxel2}, models based on SDF representations can smoothly represent complex geometry, are compact and efficient, and can be learned effectively \cite{sitzmann2019metasdf,chaudhuri2020learning}. Follow-up work has studied learning ``dual" generative models that jointly learn an uncontrained representation and one composed of primitive shapes \cite{zekun2020dualsdf,vasutalabot2022hybridsdf}.

\noindent \textbf{Implicit Representations for Physical Dynamics}. Some classical particle-based fluid simulators based on the Smoothed Particle Hydrodynamics (SPH) formalism \cite{monaghan} have transitioned from particle-based to implicit surface representations. \cite{koschier2017density,gissler2019interlinked,dfsph}. In these simulators implicit surfaces allow for smoother fluid motion with less artifacts \cite{koschier2017density}. For general-purpose learned simulators (GNNs), there is little prior work using implicit rather than particle-based representations of surfaces.

Another line of work combines implicit generative models with physical simulators \cite{mezghanni2022physical,remelli2020meshsdf}.
Mezghanni et al.~\cite{mezghanni2022physical} generate more physically stable SDF shapes by backpropagating gradients through a physical simulation layer. Their simulation mainly considers Newtonian dynamics. Our contribution is an effective, generalizable forward model for fluids that represents surfaces accurately and efficiently, and can also be used for solving inverse problems.

\section{Method}

\subsection{ 
Setup: Particle-Based Fluid Simulation}

We address the problem of learning the physical dynamics of fluids interacting with surfaces. Given the state of a physical system $X^{0}$ and some initial conditions (e.g. gravity, other external forces), rolling out a fluid simulation over $T$ steps produces a sequence of states $X^{(1)},...,X^{(T)}$. In this paper, we focus on particle-based simulation; the state of the system at time $t$ is represented by a set of particles $X^{(t)} = (p_1^{(t)},....p_n^{(t)})$, where $p_i$ is the position of particle $i$ in 3D space. A rollout over $T$ steps is produced by repeatedly applying a forward model $S_\theta: X^{(t)} \rightarrow X^{(t + 1)}$, which takes in the particle positions at time $t$ (and optionally a short history $t-1,...$) and produces the next-step particle positions.
While the forward model can be used to directly predict $X^{(t + 1)}$, most methods instead predict intermediate values $Y$, typically next-state accelerations for each particle, which can then be numerically integrated (e.g. Euler integration) to produce $X^{(t + 1)}$;
concretely, our task is to learn the parameters $\theta$ of this forward model  $f_\theta: X^{(t)} \rightarrow Y$.

\subsection{Graph Networks for Particle-Based Simulation}

Graph neural networks lend themselves naturally to particle-based simulation, since they compute pairwise interactions between nodes and aggregate these interactions.  Graph Network-based Simulators (GNS), a model architecture introduced by Sanchez-Gonzalez et al.~\cite{sanchez2020learning}, have in particular shown state-of-the-art performance on simulating fluids over long rollouts. These methods follow an ``encode-process-decode" paradigm. First, the state $X^{(t)}$ is encoded into a graph, where nodes $v_i$ are instantiated using a set of features (e.g. previous velocities, particle type, etc.); nodes $v_i, v_j$ are then connected with an edge $e_{ij}$ if their distance is less than a connectivity radius $\epsilon$. Each node and edge is associated with an embedding. The ``processor'' stage then passes the encoded graph through $M$ message-passing steps. Each message-passing step takes the current graph $\{(v_i, v_j, e_{ij})\}^{(m)}$. An edge module first computes the updated edge embeddings $e_{ij}^{(m + 1)} = f_e(v_i^{(m)}, v_j^{(m)}, e_{ij}^{(m)})$; then the edge embeddings are aggregated to compute updated node embeddings $v_i^{(m + 1)} = f_n(v_i^{(m)}, \sum_{j \in N(i)} e_{ij}^{(m)})$, where $N(i)$ are the set of neighbors of node $i$. $f_e$ and $f_n$ are generally MLPs with 2-3 hidden layers, and the weights are shared across edges/nodes (although differing across message passing steps), Finally, the ``decoder'' step applies an MLP $f_D$ to the final node embeddings to predict the accelerations $\{y_i\}$. 

In our problem, we further assume that there exists some rigid surface $S$ that interacts with the fluid throughout the simulation. The GNS approach handles surfaces by discretizing the surface into particles (i.e. a point cloud) and adding these particles $\{s_i\}$ to the graph (denoting them by a different particle type, and masking during position updates)~\cite{sanchez2020learning}. Follow-up work for mesh-based simulation handles surfaces similarly by representing the surface as an irregular mesh and adding nodes for mesh vertices~\cite{pfaff2020learning}. 

\subsection{Modeling Fluid-Surface Interactions with SDFs}
\label{sec:about_sdfs}

\begin{figure}[b!]
    \centering
    \includegraphics[width=0.30\textwidth]{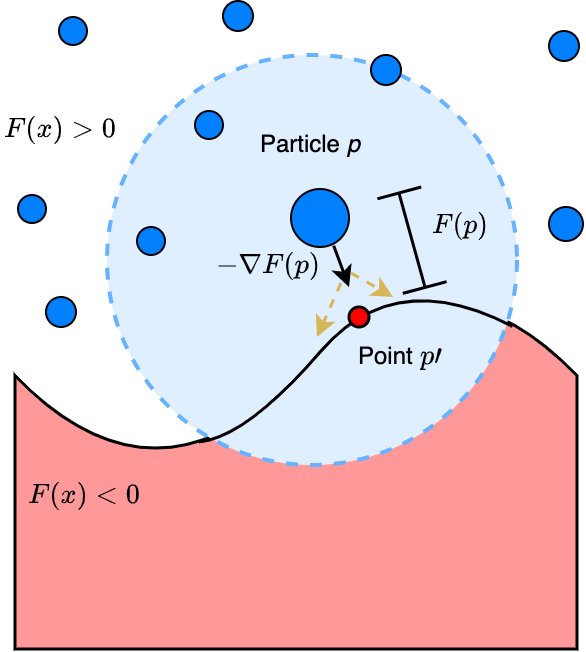}
    \caption{Rigid surfaces can be represented compactly and implicitly as the level set of a SDF: $\{x|F(x)=0\}$.
    Evaluating $F$ and its derivatives at particle position $p$ provides useful information for simulating the interaction of that particle with the surface.
    }.
    \label{fig:sdfviz_example}
\end{figure}

We propose to model the interactions of fluids and surfaces with signed distance function (SDF) representations of surfaces. An SDF represents a watertight surface as a function $F(p) = s$, which takes a spatial point $p \in \mathbb{R}^3$ and outputs its distance $s$ to the \emph{closest} corresponding point on the surface. The function is ``signed'', such that it takes on positive values for points outside the surface and negative values in the surface interior. The surface of the object is implicitly represented as the zero-level set $F(p) = 0$. Explicit representations of the surface can be constructed by applying meshing algorithms such as Marching Cubes~\cite{lorensen1987marching}.

Almost every surface can be described by an SDF. For several primitive shapes SDFs can be written down analytically (e.g. spheres, cones, cylinders); boolean operations on these shapes such as unions and intersections give rise to more complex shapes \cite{abstractionTulsiani17}. SDFs can also be represented discretely as voxel grids; or more recently for finer geometry, as neural SDFs $F_\theta(p)$ which learn to accurately approximate $F$ from mesh-based supervision.

\subsubsection{Locality and Geometric Information for Fluid-Surface Interactions}

A key insight of our method is that fluid-surface interactions are inherently about local rather than global surface geometry. The interaction of a fluid with a surface depends on a small local region of the surface, over which the geometry can have limited variation. This suggests that an effective representation of local surface geometry, trained on a variety of object shapes, can generalize 
OOD
to shapes that may look different globally.
By contrast, particles
are \emph{not} an effective representation of surface geometry; for any moderately complex surface, discretization leads to artifacts such as non-smooth fluid motion and even penetration of the surface boundary 
(we confirm this empirically).

By contrast, we argue that SDFs are a smooth, richly informative representation of local surface geometry for learning generalizable fluid-surface interactions. Our key insight is shown in Figure \ref{fig:sdfviz_example}. Given a solid surface represented as an SDF $F(p)$, we can use properties of the SDF to inform the behavior of a fluid particle as it reaches the surface. The value of the SDF $F(p)$ gives the distance from the particle at point $p$ to the closest point on the surface $p'$. Moreover, the SDF gradient $\nabla F(p)$ provides the unit-norm displacement vector from the particle to its closest surface point $\frac{(p - p')}{||p - p'||}$. This can be shown by observing that the SDF decreases most quickly on the line from $p$ to $p'$. Thus, the SDF and its gradient provide the distance and orientation of the fluid particle w.r.t. to its closest surface point; as the particle approaches the surface, its acceleration will more closely approach $\nabla F(p)$ and it will tend to rebound. Moreover, since the gradient varies smoothly because of the continuity of $F$, these features do not suffer from artifacts.

Higher order derivatives of the SDF can also be used to gain geometric information. For example, we could march to the surface by computing $p - F(x)\nabla F(x)$. At the surface, the Hessian $\nabla^2 F(x)$ is the shape operator; its eigenvalues provide the principal directions of surface curvature $\kappa_1$ and $\kappa_2$, and the mean curvature is $H = \frac{1}{2} \text{Tr}(\nabla^2 F)$. We can also estimate the surface curvature via finite differences by slightly perturbing the gradient vector on each axis and computing the SDF (Fig. \ref{fig:sdfviz_example}). In this paper, we only use the SDF and its gradient as features without curvature information; this can be seen as an \emph{effective theory} where the relevant ``local region" of the surface is a single closest point on the surface. 
As we show, modeling surfaces using SDF representations helps our fluid simulator generalize beyond the available training data by modeling fluid-surface interactions for unseen surfaces.

\subsection{Model}

\begin{figure}[]
    \centering
    \includegraphics[width=0.50\textwidth]{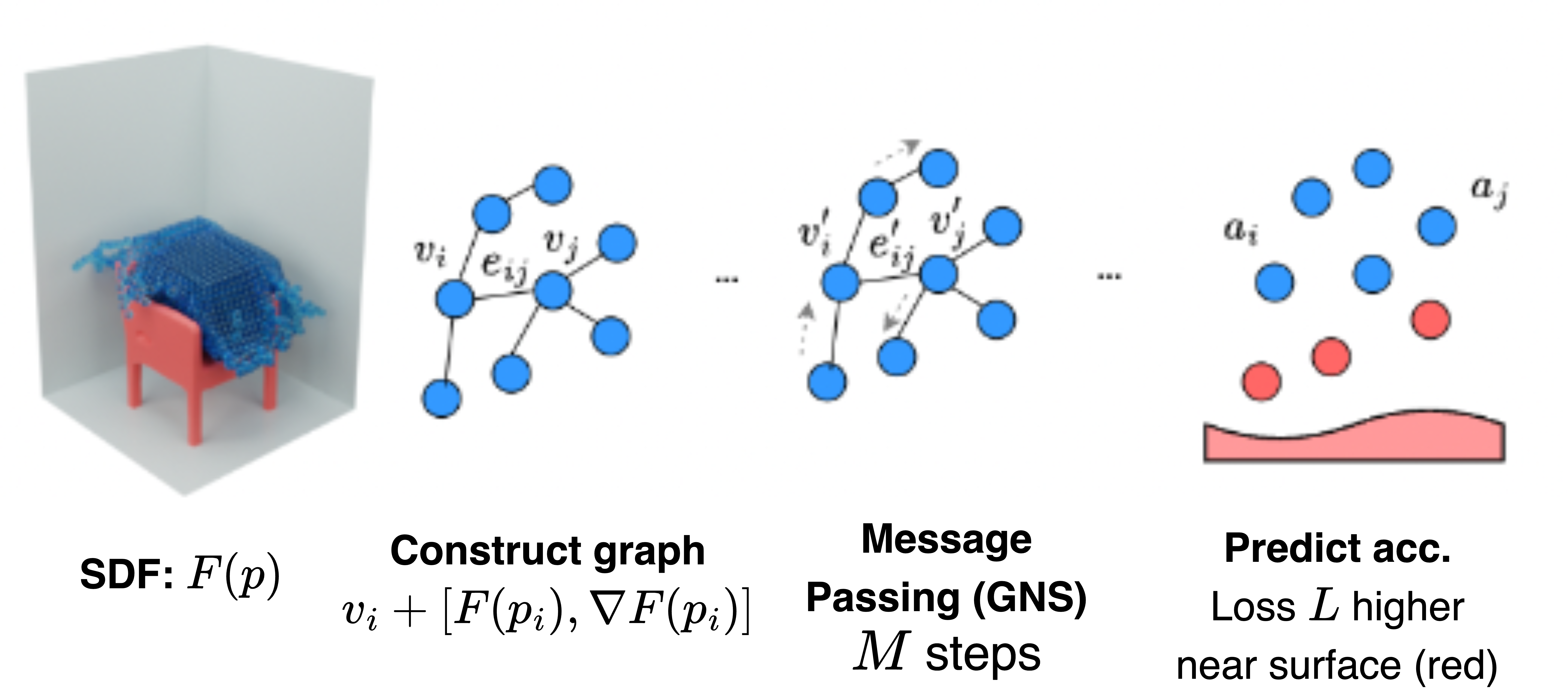}
    \caption{
    Given fluid particles and the SDF, 
    our method augments the constructed graph with features from the SDF.
    We then apply message-passing and predict dynamics,
    weighting near-surface errors more highly in our loss.
    }
    \label{fig:model_fig}
    \vspace{-2mm}
\end{figure}

Having developed our ideas about SDFs and modeling fluid-surface interactions, we now describe our approach, which we call \methodNameFull (\methodName). We learn a forward predictor ${f_\theta: X^{(t)}, F \rightarrow Y}$, which takes as input a set of particle positions and a surface implicitly parametrized by an SDF $F$. We adopt the encode-process-decode paradigm for our model, but make two key changes. In the encoder step, for each fluid particle $p_i$ we add the value of the SDF $F(p_i)$ and its gradient $\nabla F(p_i)$ to the initial node representations, such that the new representation is $v_i'^{(0)} = [v_i^{(0)}, F(p_i), \nabla F(p_i)]$. This is the way that the surface is encoded into the model; subsequently no particles representing the surface are added to the graph. The updated node embeddings are then passed through the ``processor'' with several message-passing steps and then decoded to predict the accelerations as standard. The other modification we make is to weight particles close to the surface higher during training when predicting dynamics. This has the effect of teaching the model to prioritize modeling fluid-surface interactions accurately. Since the model is predicting next-step accelerations, our loss becomes:

\begin{equation}
    L = \sum_{F(p_i) > \alpha} ||a_i^{(t)} - \hat{a_i}^{(t)}||^2 + \sum_{F(p_i) \leq \alpha} \lambda ||a_i^{(t)} - \hat{a_i}^{(t)}||^2
\end{equation}

In our experiments, we set $\alpha$ based on the ``neighborhood radius'' of the classical simulator we train against, and set $\lambda = 5$. We found that this weighting helped the model learn surface interactions more quickly without sacrificing overall accuracy on fluid prediction. Note that the SDF allows loss weighting to be implemented in a very natural way, by simple thresholding. Our approach is shown in Fig. \ref{fig:model_fig}.

\subsection{Other Advantages of SDF Representations}
\label{sec:other_advantages}
We have argued for SDFs from the perspective of achieving smooth, information-rich representations of local surface geometry. Here we mention other advantages of SDFs, each of which we demonstrate in our experimental results.

\vspace{.05in}

\noindent \textbf{Efficient message passing in the GNN}. By modeling surfaces implicitly, we reduce the number of unique particles needed to compose the 
GNN's state.
This allows us to perform more lightweight and computationally efficient simulation, as well as run the model for more message-passing steps. This advantage becomes especially clear for large objects/scenes or objects with complex surface geometry.

\vspace{.05in}

\noindent \textbf{Shape Design with SDFs}. SDFs are a useful paradigm for shape design and assembly. This is because compositions of shapes can be generated by union, intersection, and difference operations; e.g. the union of two SDFs $F_1$ and $F_2$ is $\min(F_1, F_2)$. The construction of shapes using primitives and these operations is known as Constructive Solid Geometry (CSG), and is implemented in many CAD packages for object design~\cite{foley1996computer}. Moreover the recent emergence of neural SDF models as state-of-the-art 3D shape representations, offers another compelling avenue for shape design with end-to-end learning. These suggest that good forward models with SDFs could be leveraged for design problems.  

\section{
Experiments
}

\subsection{Fluid Domains}
We evaluate our approach on complex prediction tasks involving fluids interacting with a variety of surfaces. Our main contribution here is a 3D fluid dataset consisting of water interacting with a range of primitive shapes, across shape parameters, positions, and orientations. For the ground-truth simulator we use SPlisHSPlasH \cite{benderkoschier}, a 3D-particle-based fluid simulator based on the SPH technique.


\textbf{PrimShapes Dataset}. 
We train \methodName on a 3D water dataset, where the fluid interacts with various shapes. 
We initialize a block of fluid directly above the shape, then let it fall due to gravity.
The entire scene takes place in a $1  \times 1  \times 2$ $m$ container. A random rotation is applied to the base shape which sits on the bottom of the container, and the initial position of the shape is also randomized. We extract 800 time-steps from the ground-truth fluid rollout (with $\Delta t = 0.005$); this corresponds to 4000 steps for the classical simulator, which requires very small time-steps for stability \cite{sanchez2020learning}.

We use five ``primitive'' shapes in our dataset: spheres, boxes, cylinders, cones, and toruses. These shapes have SDFs with analytical forms. (E.g. the simplest form is the origin-centered, radius-$r$ sphere with SDF $F(p) = ||p||_2 - r$; the others tend to be more complex). We emphasize that \methodName is trained \underline{only} on simulations containing these primitive shapes and generalizes zero-shot to complex objects and scenes. Our key insight is that accurate fluid-surface simulation depends on understanding local surface geometry, of which a significant diversity is expressed by these shapes (like curvature, holes, differently-sloped faces, corners). Leveraging this and the properties of SDFs, \methodName can generalize to interactions with surfaces of entirely novel \emph{global} geometry. Our training and test set consists of 1000 and 100 simulations respectively. See Supp. Fig 6 for example initializations.


\subsubsection{Generalization Test-Sets}
We create a range of qualitative and quantitative evaluation scenarios to test our model's generalization ability. The first three below assess generalization to novel analytical SDFs (e.g. unions $\min(f_1, f_2)$ or differences $\max(f_1, -f_2)$), and the last to real-world shapes represented as neural SDFs $f_\theta$.

\vspace{3mm}

\noindent \textbf{Prim-OOD} (60 simulations). This contains the same shapes with 
OOD
shape parameters from the training set (e.g. cylinder with higher radius $r$ and lower height $h$).

 \noindent \textbf{Prim-Unions} (40). This test set includes all possible pairwise unions which `merge' two of the primitive shapes. See Supp. for examples.
 
 \noindent \textbf{Funnels} (12). We examine the model's ability to handle analytical ``differences" of shapes by creating funnels, formed by carving an inverted cone out of a cylinder.

\noindent \textbf{`Complex' Scenes} (7). As the key, challenging test of our model's generalization, we simulate fluid interacting with several complex real-world objects. To do so, we leverage neural SDFs of shapes, which can represent complex shapes with high accuracy and fidelity (Sec. \ref{sec:about_sdfs}). Instead of an analytical SDF $f(p)$, we use a trained neural SDF $f_\theta(p)$ to generate our SDF features during our forward pass. To test our model's ability to generalize zero-shot to complex objects/scenes, we train a single SIREN model \cite{sitzmann2019siren} for 7 fine-grained objects (including the coral, lion, and room scenes shown in \cref{fig:teaser}, \cref{fig:main_rollouts}). For each object we create a fluid scene using the ground-truth simulator, and evaluate \methodName on this small \textbf{Complex-Scenes} dataset. To represent the geometry at an appropriate scale, these objects are scaled $2$x in the fluid scene compared to previous datasets. To test our model on \textit{families} of shapes, which gives us the capacity to solve inverse problems, we train a variational DeepSDF model \cite{park2019deepsdf} for the ShapeNet chair and bowl categories. See Supp. for details on these scenes and models.

\begin{SCtable*}[1][h!]
\centering
\small
\setlength{\tabcolsep}{3pt}
\begin{tabular}{l|ll|ll|ll|ll}
\toprule
            & \multicolumn{2}{c|}{Chamfer} & \multicolumn{2}{c|}{Chamfer Surface} & \multicolumn{2}{c|}{Number Inside} & \multicolumn{2}{c}{Mean SDF Inside} \\
Testing Set & \methodName & GNS \cite{sanchez2020learning} & \methodName & GNS & \methodName & GNS & \methodName & GNS \\
\midrule
Primitives & 0.0297 & 0.0285 & 4.704 & 5.451 & 0 & 2000 & -1.04e-11 & -2.48e-05\\
Primitives-OOD & 0.0321 & 0.0343 & 10.951 & 18.445 & 324 & 33006 & -1.65e-06 & -0.0006 \\
Primitives-Unions & 0.0380 & 0.0348 & 24.553 & 26.307 & 10000 & 63000 & -4.46e-05 & -0.001 \\
Funnels & 0.0362 & 0.0521 & 9.990 & 13.663 & 2 & 7204 & -2.51e-09 & -0.001\\
Complex-Scenes & 0.0638 & 0.2658 & 50.521 & 276.176 & 40513 & 257722 & -0.0011 & -0.026 \\

\bottomrule
\end{tabular}
\caption{Results on in-distribution (Primitives) and OOD test sets, comparing \methodName to GNS \cite{sanchez2020learning} which represents the surface as particles.}
\label{tab:main_results}
\end{SCtable*}

\subsection{Training Details}

For all experiments, we train our model on the PrimShapes dataset. The training data is pairs $X^{(t)}, X^{(t + 1)}$ from the ground-truth simulator. We use a graph network (GN) architecture with 10 message-passing steps and train with Adam for 2M steps (see Supp. for details).

\subsection{Metrics and Baseline}
\label{sec:metrics}
Although we train on next-step prediction, we measure performance on full rollouts. We use mean Chamfer distance between the predicted particles $P$ and ground-truth $G$ to measure overall rollout agreement (averaging across timesteps). Compared to MSE, Chamfer distance is permutation-invariant and therefore a better measure of similarity between fluid flows \cite{sanchez2020learning}. This metric, however, can be noisy since it averages over long rollouts on chaotic systems with many particles. Since we focus on the quality of fluid-surface interactions, we also compute a ``Chamfer surface" metric that focuses on prediction errors near the surface. For each time-step, we consider the predicted point cloud only near the surface (where $F(p) < \alpha$ for threshold $\alpha$) and measure the Chamfer distance of $P_{\alpha}$ to $G$, summing over particles; we do this symmetrically for $G$. Intuitively this focuses on discrepancies between $P$ and $G$ for fluid motion near the surface. See Supp. for further details.

Since surface penetration is a serious failure mode, we also measure the number of particles inside the surface through the rollout (summing over the number of time-steps). Additionally, we compute the average SDF value for particles \emph{inside} the surface, indicating the extent of surface penetration (since SDFs become increasingly negative deeper into the surface interior). 

\noindent \textbf{Baseline}. Our main comparison is to the graph network simulator (GNS) proposed by Sanchez Gonzalez et al. \cite{sanchez2020learning}. Their approach represents surfaces explicitly through discretization into particles, which are added to the graph and participate in the message-passing. To implement this baseline, we convert all the SDFs in our datasets to meshes using Marching Cubes \cite{lorensen1987marching}. From this mesh we uniformly sample $2000$ surface particles (comparable to the number of fluid particles in our simulations) and add them to the graph. In all our experiments we compare our implicit \methodName approach to this baseline with explicit surface particles.

While the baseline method could in principle be extended to include local surface information (e.g. using surface normal estimates for each particle as additional features, or mesh connectivity), we omit such an extension here.
This is because any explicit representation of large surfaces quickly becomes computationally infeasible, especially for large objects/scenes or complex geometry. Whereas the simulation quality of \methodName scales with the capacity of the SDF network $f_\theta$, the baseline quality scales with the number of particles.
We show in Sec \ref{sec:results} that \methodName still outperforms (with lower computational cost and inference time) a baseline with more particles.

\section{
Results
}
\label{sec:results}

\begin{figure*}
        \centering
        \includegraphics[width=\linewidth]{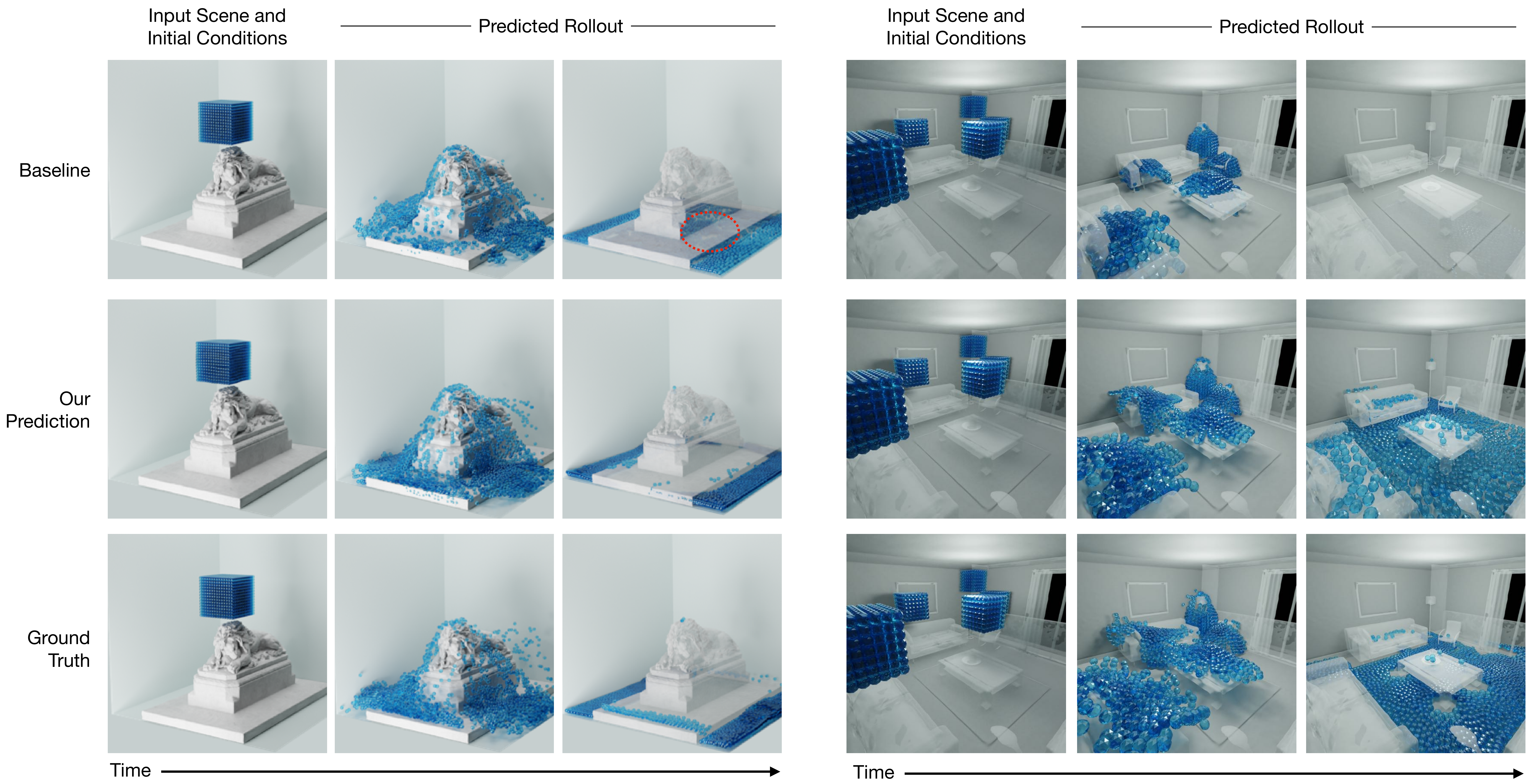}
        \caption{ 
        Although \methodName is only trained on primitive shapes, our model generalizes to complex and real-world shapes without any additional training. We visualize our predictions (middle) for two scenes. In comparison, the baseline (top) often generates unrealistic predictions, such as having water penetrate solid objects. In many cases, the predictions from the baseline cause the water to disappear because it falls through the floor (top right).} 
        \label{fig:main_rollouts}
    \end{figure*}

\vspace{-0.1cm}
\methodName can accurately predict fluid behavior over long rollouts, interacting with a range of complex surfaces (Fig. \ref{fig:main_rollouts}). Across the different analytical primitive shapes in our test sets and different orientations, we can generate accurate and visually realistic predictions of fluid behavior. See the Supp. Material for \textbf{videos} of these, and all, rollouts in the paper. Our \emph{key result} is the following: we find that \methodName can perform effective simulation with surfaces that are significantly out-of-distribution, and even \underline{generalize to complex real-world shapes and scenes}. This demonstrates that \methodName can leverage the rich geometric information provided by the SDF for generalizing to novel surfaces expressed as neural SDFs.

\noindent \textbf{Quantitative Results and Comparison to Baseline}. We evaluate \methodName on rollouts from ID and OOD test sets of shapes with analytical SDFs, and on complex scenes, with comparison to the GNS baseline. Results are shown in Table \ref{tab:main_results}. Across all test sets, our model's fluid predictions display significantly lower surface penetration than the baseline (e.g. 324 vs. 33006 particles on average for the primitives+OOD test set). Moreover, for those few particles which \emph{have} penetrated the surface, the average SDF value for our method is nearly zero; this implies that any particle that penetrates the boundary ($F(p) < 0$) is quickly projected outward to the region $F(p) > 0$. On Chamfer distance near the surface, which measures \emph{overall} accuracy of fluid-surface interactions beyond surface penetration, our method consistently improves over the baseline. Overall Chamfer distance between the two models is generally comparable. This is unsurprising given that small differences in particle positions along long time-horizons, potentially far from the surface, can accumulate error.
(\cref{sec:metrics}). See Supp. Fig. 5 for rollouts on rows 1-4 of Table \ref{tab:main_results}, indicating the stability and accuracy of our method's dynamics. 





\begin{figure}[t]
    \centering
    \includegraphics[
    width=0.40\textwidth
    ]{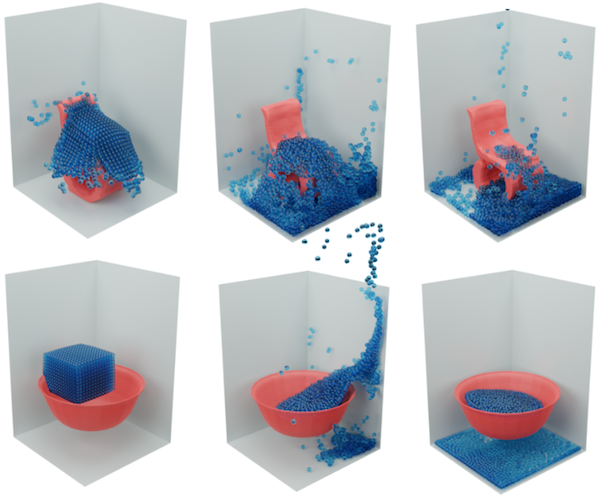}
    \caption{
    \methodName predictions on ShapeNet objects.
    }
    \label{fig:chair_rollouts}
    \vspace{-0.1cm}
\end{figure}

\noindent \textbf{Generalizing to Complex Scenes}
We evaluate \methodName on chairs and bowls from Shapenet, and fine-grained objects and scenes in the real world (See \cref{fig:teaser}, \cref{fig:main_rollouts}, \cref{fig:chair_rollouts}, and \cref{fig:mountain}, also videos in Supp.). Remarkably \methodName trained only with primitive shapes generalizes zero-shot to complex shapes/scenes represented as neural SDFs. This is notable because our chosen real-world objects differ \emph{widely} (see the variation in \cref{fig:main_rollouts} and \cref{fig:chair_rollouts}), and second, neural SDFs are imperfect compared to analytical SDFs (e.g. not fulfilling the property $||\nabla f|| = 1$). Despite these challenges, \methodName produces rollouts with accurate and realistic fluid dynamics, capturing fine-grained behavior near complex surfaces. Our method also avoids unintuitive artifacts, such as surface penetration, seen in the GNS baseline. 

\begin{figure}
    \centering
    \includegraphics[width=0.50\textwidth]{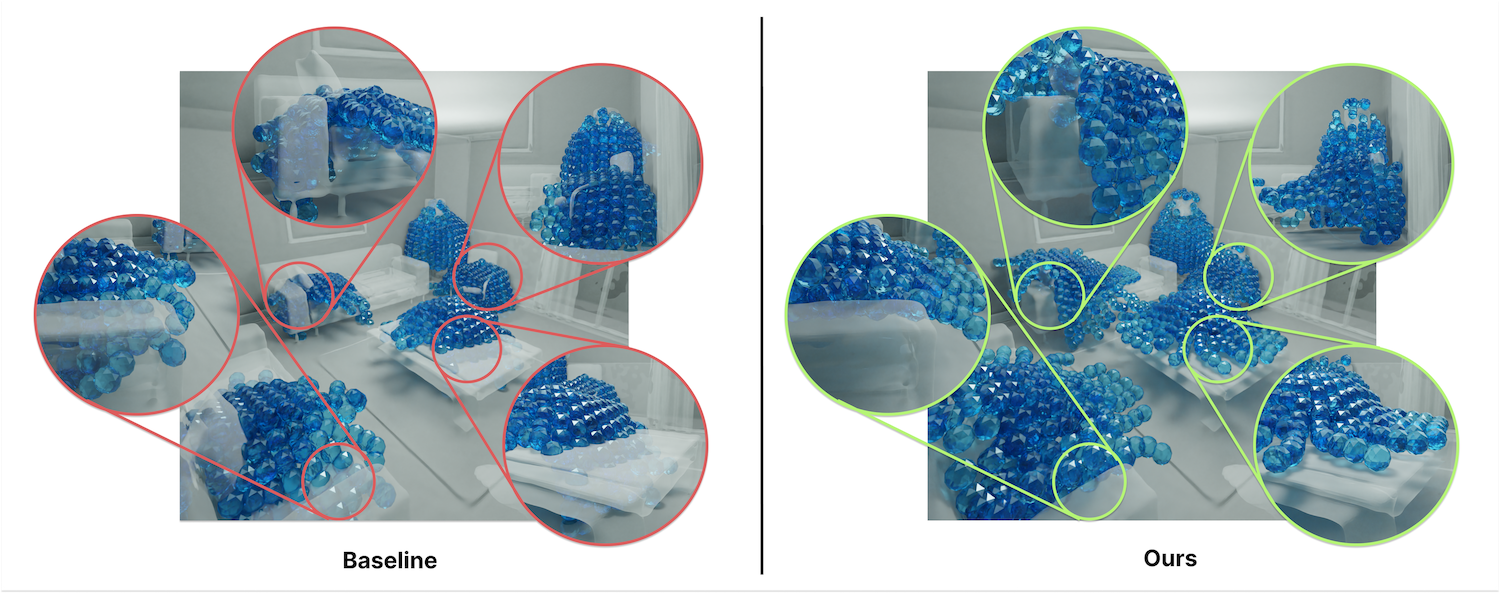}
    \caption{Fine-grained comparison of \methodName and GNS on four parts of the room scene. Our dynamics accurately model the furniture while the baseline's leak into surfaces.}
    \label{fig:room_penetration}
    \vspace{-4mm}
\end{figure}

Quantitative results (Table \ref{tab:main_results}, row `Complex-Scenes') indicate that our method outperforms the baseline most significantly on complex real-world scenes. These results elucidate key generalization advantages of our method in handling surfaces with complex geometry and large scenes at scale. For the coral scene (\cref{fig:teaser}), our method's fluid rollouts accurately capture and percolate through the fine tentacle structure of the coral. The GNS baseline does not display this ability (Supp. Fig 1). Our method also scales accurately to large scenes such as the room scene in \cref{fig:teaser}. By contrast, the baseline's predictions consistently exhibit penetration of the furniture surfaces in the scene due to insufficient resolution (see \cref{fig:room_penetration}). Recall (Sec. \ref{sec:metrics}) that our scaling law is the capacity of the neural SDF network $f_\theta$, instead of the number of surface particles. 



\begin{figure}[!ht]
    \centering
    \includegraphics[width=0.50\textwidth]{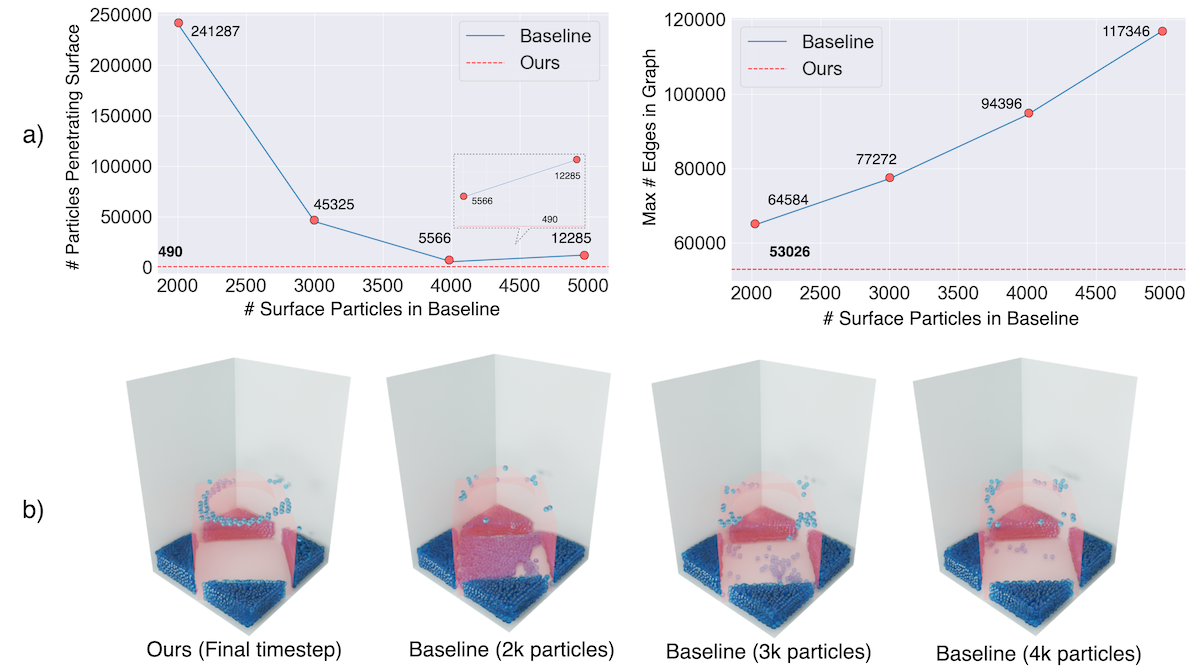}
    \caption{Comparing our method vs. the baseline with more surface particles. (a) Left shows how surface penetration is affected; even as the baseline improves with more surface particles, our method maintains an order magnitude improvement (red line, 490 particles). Right shows that the graph size increases significantly with the number of surface particles; we measure the max number of edges during the rollout, which indicates memory cost. (b) Final rollout timesteps for our method and the GNS at 2000, 3000, and 4000 particles, which reflect the quantitative results.}
    \label{fig:particle_analysis}
    \vspace{-1mm}
\end{figure}

\begin{figure}
    \centering
    \includegraphics[width=0.50\textwidth]{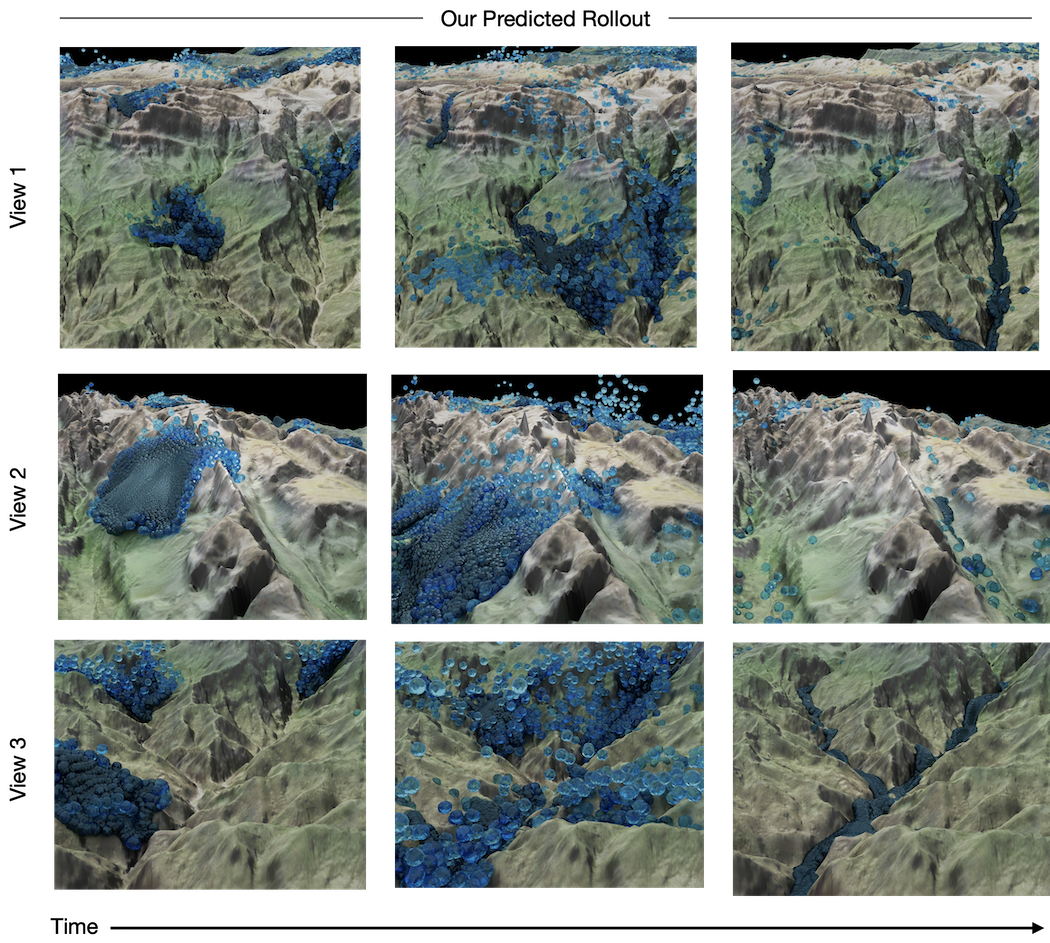}
    \caption{Predicted fluid rollouts from \methodName on different parts of a $40 \times 27 \times 6$ $m^3$ mountain scene. }
    \label{fig:mountain}
    \vspace{-5mm}
\end{figure}

\noindent \textbf{Efficiency of Simulation}. We study comparing \methodName as we increase the number of surface particles for the GNS baseline (which in our experiments is afforded 2000 surface particles). Note that there is a tradeoff between performance and computational cost (memory, time) for the baseline; more particles improves surface resolution while also increasing graph sizes. We examine a scene in the Primitives-Unions test set (union of a sphere and box), for which GNS's predictions display significant surface penetration, and increase surface particles from 2000 to 5000. Correspondingly GNS's fluid predictions improve, with less surface penetration (\cref{fig:particle_analysis}, top left and bottom). However, the baseline with 5000 particles still shows an \emph{order magnitude more} penetration than our method (12285 vs. 490 particles). Moreover its computational cost, measured as the maximum particle graph size (\# edges) during the rollout, almost doubles with 5000 surface particles (\cref{fig:particle_analysis}, top right). In contrast \methodName achieves near-zero surface-penetration with constant, smaller fluid graphs, overcoming the baseline's cost-performance tradeoff. We similarly analyze the \emph{room scene} (\cref{fig:teaser}), increasing the number of surface particles to 5K and 10K. Remarkably, \methodName here outperforms even a \underline{10000}-particle baseline (Ours 125.39 vs. baseline 194.98 Chamfer Surface; Supp.for details). This suggests fundamental advantages in scaling for simulation with implicit surfaces. We perform an inference time study (Supp.) showing that our method achieves speedups over GNS and the ground-truth simulator.

\textbf{Scaling up \methodName}. As a final benchmark of unusual size and complexity, we assessed \methodName on a natural scene with complex topography (Puncak Jaya mountains, Indonesia). This scene was $40 \times 27 \times 6$ $m^3$ in size ($6840$ $m^3$ compared to roughly $10$ $m^3$  for previous scenes), and contained 68,600 fluid particles (order magnitude more than previous scenes, see Supp. for details). \cref{fig:mountain} shows the predictions of \methodName, which correctly handle complex features such as slopes, confluences, and valleys between the mountains. Note that while GNS would require $O(1m)$ particles to represent this scene, we leave the network size for SIREN \emph{unchanged} from smaller scenes. These results demonstrate the potential of our method for large-scale, realistic dynamics. Please see Supp. for rollout videos.

 
\noindent \textbf{Inverse Design}. Finally, we evaluate \methodName on its potential to solve object design tasks. We show the ability of our model to design both parametrized shapes, by optimizing the shape parameters with gradients, as well as shapes represented by neural SDF models. 

\noindent \textit{Parametrized Design Tasks}. We consider the funnel, which is an inverted capped cone carved out of a cylinder, for which we have derived an analytical SDF. This shape is parametrized by the cylinder radius $R$, height $H$, cone height $h = H$; and the radii $r_1$ and $r_2$ of the bottom and top circles that ``cap'' the cone, with  $r_1 > r_2$. 

In our setup, we drop a block of fluid directly onto the object. We consider two tasks: designing a \emph{bowl} where the task is to contain the fluid inside the object, and a \emph{funnel} where the task is to concentrate the water onto a location onto the ground. We optimize $r_1$ (which we call the ``capture radius'' that captures fluid into the object) and $r_2$ (the ``filter radius'' which decides how much water to filter onto the ground). For the funnel, the desired solution is large $r_1$ and small $r_2$, and for the bowl we want $r_2 \rightarrow 0$. If our learned model is performing well, the model should be able to discover these solutions.

We roll out our model for 50 (bowl) and 75 (funnel) timesteps. Our reward measures the log probability of a Gaussian centered at the desired location of the fluid particles. For the bowl task, we use the bottom of the object, i.e. $\mathcal{N}((0., 0., h), \Sigma)$, where $h$ is the object height. For the funnel we use a 2D Gaussian $\mathcal{N}((0., 0.), \Sigma)$ and measure the reward on particles that reach the ground during training. For a given design, we roll out our model and obtain gradients, which we use to iteratively refine the design. 

\begin{figure}[t]
    \centering
    \includegraphics[width=0.48\textwidth]{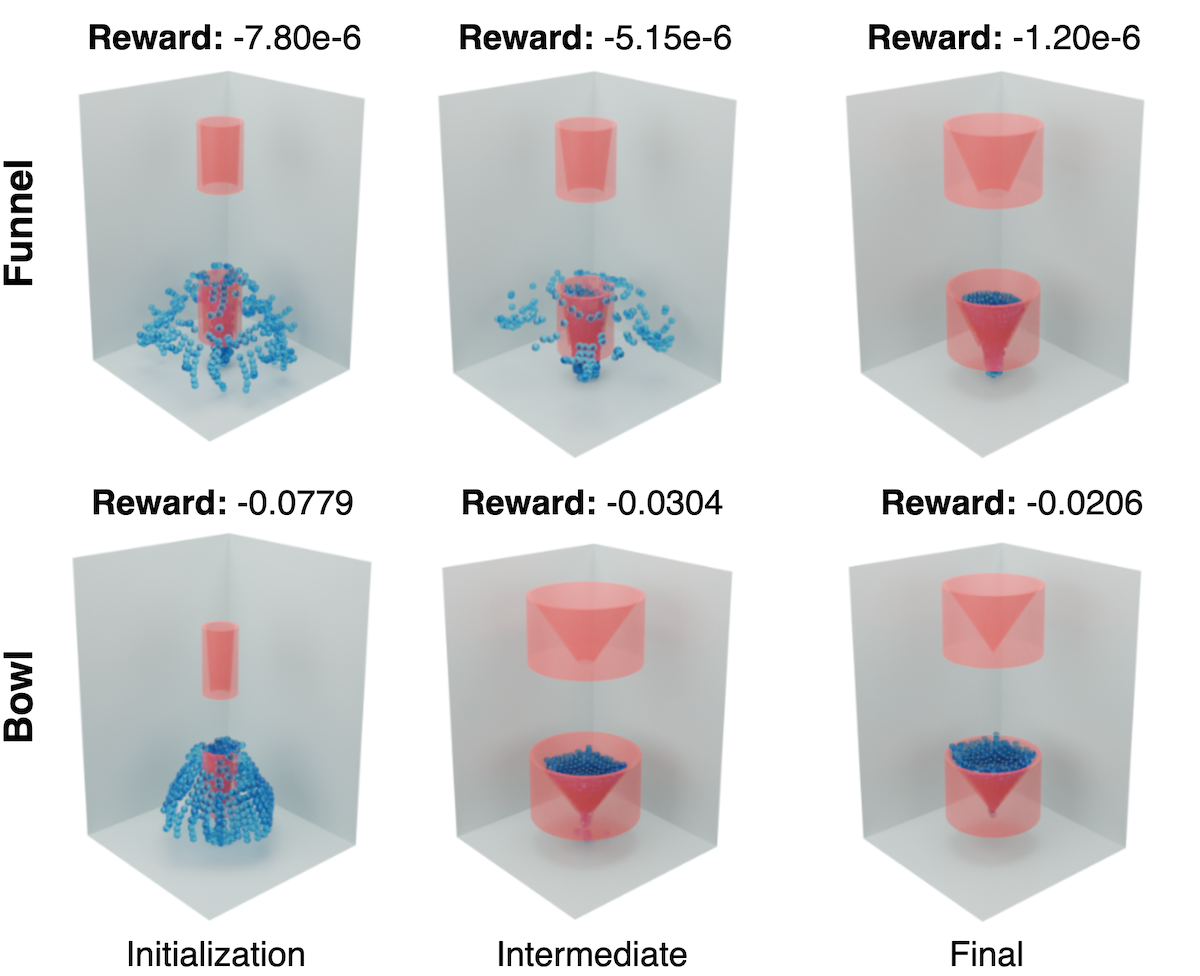}
    \caption{Design optimization; top is the funnel task, bottom the bowl task. For each frame, the design is shown near the top of the container and a timestep from the fluid rollout is shown below. Initial, intermediate, and final designs are shown; note that the reward increases through optimization and final designs approximate a funnel and bowl well.}
    \label{fig:design}
\end{figure}

Qualitative results are shown in Fig. \ref{fig:design}. Starting from a sub-optimal design with low $r_1$ and $r_2 > 0$, our model successfully converges for the bowl; similarly the funnel converges from a suboptimal to an optimal solution. This shows that over long rollouts and hundreds of message-passing steps with dense fluid-surface interactions, our model's gradients are useful for solving inverse design problems. We also find that design is robust across different initializations of $r_1$ and $r_2$ (details in Supp.).

\vspace{3mm}

\begin{figure}[t]
    \centering
    \includegraphics[width=0.49\textwidth]{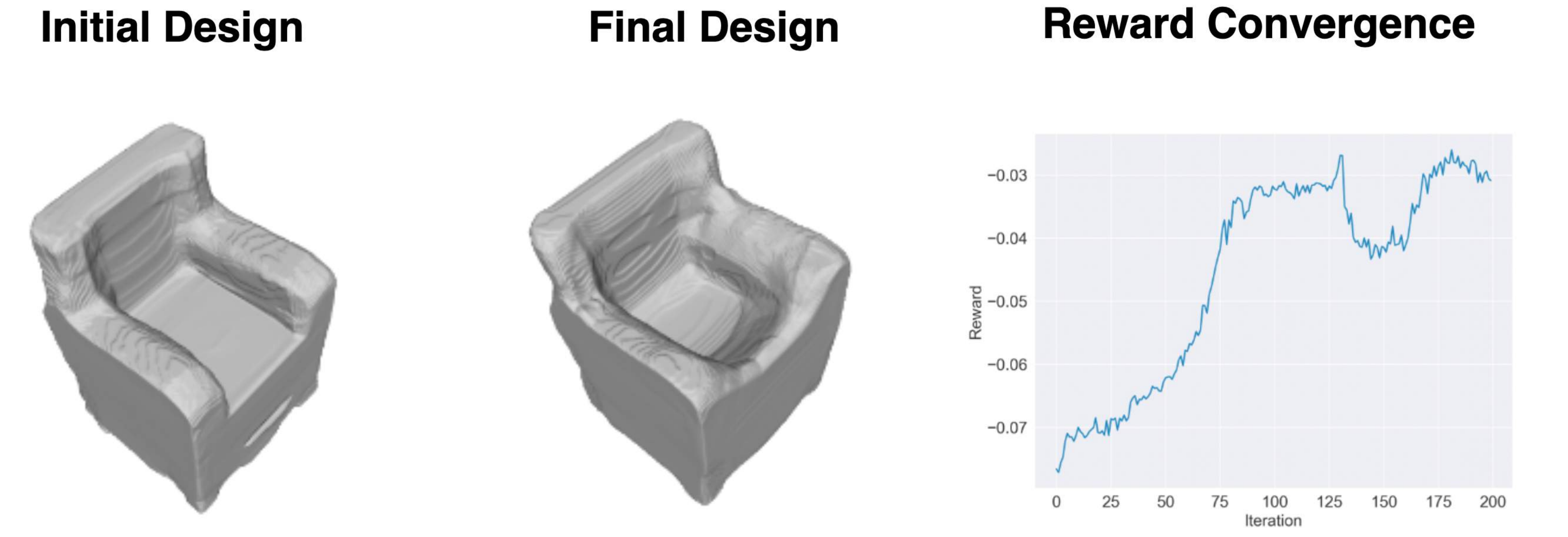}
    \caption{
    We differentiate through \methodName to directly optimize (in the latent space of a DeepSDF) a chair that holds water dropped onto it.
    }
    \label{fig:chair_design}
    \vspace{-6mm}
\end{figure}

\noindent \textit{Latent Space Design}. We further examined whether \methodName could directly optimize in the latent space of a DeepSDF model, trained on ShapeNet chairs. For this investigation, as a proof-of-concept we aimed to discover a novel chair that traps and contains falling water (i.e. maximizes the ``bowl reward'' above). Given a DeepSDF model $f_\theta(p, z)$ with latent code $z_0$, we iteratively refine the design by rolling out our simulator for 100 time-steps, computing a reward, and updating $z$ via gradient-based optimization. 
Across multiple trials, our model converges from an suboptimal initial chair to a final design that can contain water and retains the semantics of a chair~(Fig. \ref{fig:chair_design}).

\section{Conclusion}
In this paper, we have taken a first step towards integrating implicit 3D representations into the dynamical simulation of physical models. \methodName realizes this approach by incorporating the implicit representation of solid surfaces (via SDFs) into a graph neural network fluid simulator.
As such, \methodName is able to accurately model how fluids interact with surfaces, including highly complex surfaces unseen in training.
We find that promoting this type of OOD generalization is helpful in both forward and inverse contexts.

{\small \textbf{Acknowledgements}: We thank Huy Ha for help with visualizations of fluid rollouts. We thank Jan Bender for helpful feedback on SPlisHSPlasH, and Kelsey Allen, Tobias Pfaff, 
and Alvaro Sanchez-Gonzalez for advice on graph
network simulators. We thank Max Helman for early setup work on SPlisHSPlasH. This research is partially supported by the NSF STC for Learning the Earth
with Artificial Intelligence and Physics, and the NSF NRI Award
\#1925157. AM is supported by the NSF fellowship.}


 
{\small
\bibliographystyle{ieee_fullname}
\bibliography{egbib}
}

\clearpage

\end{document}